\documentclass[11pt]{article}

\usepackage[final]{acl}

\usepackage{times}
\usepackage{latexsym}

\usepackage[T1]{fontenc}

\usepackage[utf8]{inputenc}

\usepackage{microtype}

\usepackage{inconsolata}

\usepackage{graphicx}

\usepackage{amsmath}
\usepackage{enumitem}

\usepackage{tikz}
\usetikzlibrary{arrows.meta,positioning,shapes.geometric,fit,backgrounds,patterns,matrix,calc}
\usepackage{booktabs}
\usepackage{multirow}
\usepackage{makecell}
\usepackage{graphicx}
\usepackage{array}
\newcolumntype{L}[1]{>{\raggedright\arraybackslash}m{#1}}
\newcolumntype{C}[1]{>{\centering\arraybackslash}m{#1}}
\newcommand{\lcell}[1]{%
  \begin{minipage}[c]{\linewidth}
    \raggedright
    #1
  \end{minipage}%
}

\newcommand{\ccell}[1]{%
  \begin{minipage}[c]{\linewidth}
    \centering
    #1
  \end{minipage}%
}

\newcommand{\errcell}[1]{%
  \raisebox{-0.55\height}{\parbox{\linewidth}{\centering #1}}%
}
\usepackage[most]{tcolorbox}
\usepackage{xcolor}
\usepackage{listings}
\usepackage{booktabs}
\usepackage{array}
\usepackage{multirow}
\usepackage{graphicx}
\usepackage{stfloats}
\usepackage{placeins}
\usepackage{adjustbox}
\title{MASTE: A Multi-Agent Pipeline for \\Zero-Shot Aspect Sentiment Triplet Extraction}

\author{
 \textbf{Ao Hong\textsuperscript{1}},
 \textbf{Lehang Wang\textsuperscript{2}},
 \textbf{Zhirun Yue\textsuperscript{1}},
 \textbf{Mingxin Wang\textsuperscript{1}},
 \textbf{Zihan Wang\textsuperscript{1}},
 \textbf{Houde Liu\textsuperscript{1}\thanks{Corresponding author.}}
\\
\\
 \textsuperscript{1}Tsinghua University,
 \textsuperscript{2}Wuhan University,
 }

\begin{document}
\maketitle

\begin{abstract}
Aspect Sentiment Triplet Extraction (ASTE) requires jointly identifying
\emph{(aspect, opinion, sentiment)} triples from a given review sentence.
While large language models (LLMs) achieve strong zero-shot performance on many NLP benchmarks, their effectiveness on ASTE remains limited, as single-pass generation forces the model to commit to span boundaries, opinion grouping, and polarity in one decoding step.
Common remedies---few-shot in-context learning and chain-of-thought prompting---offer only marginal improvements and rely heavily on either in-domain demonstrations sampled from labeled training data or carefully engineered reasoning prompts,
neither of which is broadly available in zero-shot deployment. 
Inspired by the classical agent paradigm, we propose
\textbf{MASTE} (\textbf{M}ulti-\textbf{A}gent pipeline for zero-shot
\textbf{A}spect \textbf{S}entiment \textbf{T}riplet \textbf{E}xtraction),
a four-stage framework in which specialized agents handle each compositional subtask sequentially with explicit conditioning on prior outputs.
This design enables entirely training-free zero-shot ASTE and generalizes across different backbones and datasets.
Extensive experiments on four ASTE benchmarks show that MASTE substantially outperforms zero-shot and chain-of-thought LLM baselines under the same backbone, narrowing the gap to fully supervised methods without using any labeled triplets.
Our code is available at \url{https://github.com/Hankerlove/MASTE}.
\end{abstract}

\section{Introduction}

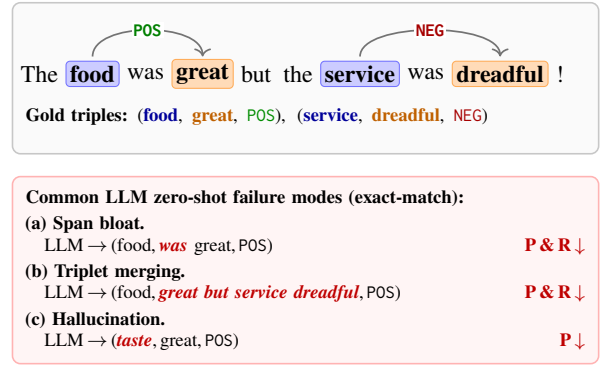
\begin{figure}[t]
\centering
\begingroup
\def\badspan#1{{\color{red!75!black}\itshape\bfseries #1}}
\def\errEffect#1{{\scriptsize\bfseries\color{red!75!black}#1}}

\begin{tikzpicture}[
  tok/.style    ={font=\footnotesize, inner sep=1.6pt, anchor=base},
  asp/.style    ={tok, fill=blue!20,   draw=blue!55,   rounded corners=1.5pt,
                  font=\footnotesize\bfseries},
  opn/.style    ={tok, fill=orange!28, draw=orange!65, rounded corners=1.5pt,
                  font=\footnotesize\bfseries},
  arc/.style    ={->, line width=0.65pt, draw=black!55,
                  shorten <=1pt, shorten >=1pt},
  poslab/.style ={font=\scriptsize\bfseries, text=green!50!black,
                  fill=white, inner sep=0.6pt},
  neglab/.style ={font=\scriptsize\bfseries, text=red!65!black,
                  fill=white, inner sep=0.6pt},
  panel/.style  ={rectangle, rounded corners=3pt, inner sep=5pt, align=left,
                  line width=0.6pt},
]

\node[tok]                      (w1)  at (0,0) {The};
\node[asp, right=1pt of w1]     (w2)              {food};
\node[tok, right=1pt of w2]     (w3)              {was};
\node[opn, right=1pt of w3]     (w4)              {great};
\node[tok, right=1pt of w4]     (w5)              {but};
\node[tok, right=1pt of w5]     (w6)              {the};
\node[asp, right=1pt of w6]     (w7)              {service};
\node[tok, right=1pt of w7]     (w8)              {was};
\node[opn, right=1pt of w8]     (w9)              {dreadful};
\node[tok, right=1pt of w9]     (w10)             {!};

\coordinate (arcBase) at ($(w1.base)+(0,0.30cm)$);
\coordinate (labY)    at ($(w1.base)+(0,0.68cm)$);
\coordinate (arcTop)  at ($(w1.base)+(0,0.80cm)$);

\coordinate (foodA)  at (w2.north |- arcBase);
\coordinate (greatA) at (w4.north |- arcBase);
\coordinate (servA)  at (w7.north |- arcBase);
\coordinate (dreadA) at (w9.north |- arcBase);

\draw[arc]
  (foodA) .. controls (foodA |- arcTop) and (greatA |- arcTop) .. (greatA);

\draw[arc]
  (servA) .. controls (servA |- arcTop) and (dreadA |- arcTop) .. (dreadA);

\coordinate (posX) at ($(foodA)!0.5!(greatA)$);
\coordinate (negX) at ($(servA)!0.5!(dreadA)$);

\node[poslab] (posL) at (posX |- labY) {\texttt{POS}};
\node[neglab] (negL) at (negX |- labY) {\texttt{NEG}};

\node[inner sep=0pt, fit=(w1)(w10)(posL)(negL)] (sentarea) {};

\node[font=\scriptsize, below=4pt of sentarea.south west, anchor=north west,
      text width=7.4cm, align=left]
  (out) {%
   \textbf{Gold triples:}
   ({\color{blue!60!black}\textbf{food}},\,
    {\color{orange!75!black}\textbf{great}},\,
    {\color{green!55!black}\texttt{POS}}),\;
   ({\color{blue!60!black}\textbf{service}},\,
    {\color{orange!75!black}\textbf{dreadful}},\,
    {\color{red!65!black}\texttt{NEG}})};

\begin{scope}[on background layer]
    \node[panel, fill=black!2, draw=black!22,
          fit=(sentarea)(out)(arcTop),
          inner xsep=5pt, inner ysep=7pt, text width=7.4cm] (topbox) {};
\end{scope}

\node[panel, fill=red!4, draw=red!30, below=8pt of topbox.south west,
      anchor=north west, text width=7.4cm, font=\scriptsize]
  (fail) {%
\textbf{Common LLM zero-shot failure modes (exact-match):}
\\[2pt]
\textbf{(a) Span bloat.}\\
\quad LLM\,$\to$\,(food,\,\badspan{was} great,\,\texttt{POS})
\hfill\errEffect{P\,\&\,R\,$\downarrow$}\\[2pt]
\textbf{(b) Triplet merging.}\\
\quad LLM\,$\to$\,(food,\,\badspan{great but service dreadful},\,\texttt{POS})
\hfill\errEffect{P\,\&\,R\,$\downarrow$}\\[2pt]
\textbf{(c) Hallucination.}\\
\quad LLM\,$\to$\,(\badspan{taste},\,great,\,\texttt{POS})
\hfill\errEffect{P\,$\downarrow$}};

\end{tikzpicture}
\endgroup

\caption{(\textit{Top}) An illustration of the ASTE task:
given an input sentence, the model is required to extract the complete
set of \texttt{\textbf{(Aspect, Opinion, Polarity)}} triples.
{\color{blue!60!black}\texttt{\textbf{Aspects}}} (blue) and
\texttt{\color{orange!75!black}\textbf{opinions}} (orange) are paired by arcs
whose labels denote the sentiment \texttt{\textbf{polarity}}.
(\textit{Bottom}) Three possible systematic failure modes that zero-shot LLM prompting
exhibits on the same example; each impairs Precision~(\textbf{P}),
Recall~(\textbf{R}), or both.}
\label{fig:task}
\end{figure}

\textbf{Aspect-Based Sentiment Analysis (ABSA)} is a fine-grained sentiment analysis task that identifies the sentiment polarity expressed toward each specific aspect mentioned in a given text~\citep{pontiki2014semeval}.
Among its subtasks, \textbf{Aspect Sentiment Triplet Extraction (ASTE)}
~\citep{peng2020knowing} is the most integrated formulation. 
As shown in Figure~\ref{fig:task} (\textit{Top}),  given a review sentence, ASTE returns the complete set of \texttt{(Aspect, Opinion, Polarity)} triples, where the \texttt{Aspect} is the term being evaluated, \texttt{Opinion} is the expression that describes the aspect, and \texttt{Polarity} denotes the corresponding sentiment polarity, including positive, negative, and neutral.
For example, given the review sentence \emph{``The food was great but the
service was dreadful\,!''}, an ASTE system is expected to output both
$(\textit{food},\,\textit{great},\,\texttt{POS})$ and
$(\textit{service},\,\textit{dreadful},\,\texttt{NEG})$.

Previous approaches to ASTE and related ABSA subtasks fall into several categories, including pipeline-based methods~\citep{peng2020knowing}, sequence tagging~\citep{xu2020position,yan2021unified}, sequence-to-sequence
generation~\citep{zhang2021towards,naglik2024aste},
grid and table filling tagging schemes~\citep{wu2020grid,chen2022enhanced,sun2024miniconGTS}, and
multi-domain joint training~\citep{hou2024trainonce}.
Despite their strong in-domain performance, these supervised methods share
two limitations: (i) they depend on dataset-specific triplet
annotations that limit cross-domain transfer, and (ii) they are bound to a fixed annotation convention (e.g., span-boundary style) and require re-training when transferred to new domains or extraction granularities.

Recently, large language models (LLMs) have demonstrated remarkable performance and strong zero/few-shot generalization across a wide range of NLP tasks~\citep{brown2020gpt3,deepseekv3}, including ABSA~\citep{zhang2024sentiment,scaria2024instructabsa,yang2024faima,fan2025aspect}.
However, when applied directly to ASTE, LLMs struggle with this
fine-grained structured task: they often fail to capture the complex latent
dependencies between aspect and opinion terms, confuse aspect mentions with
opinion mentions, and produce systematic span-boundary errors and hallucinated spans.
Concretely, even GPT-4o (zero-shot) reaches only
$35.4\%$ F1 score on \textsc{14res}, leaving a roughly $40$-point gap to state-of-the-art supervised methods~\citep{sun2024miniconGTS}.
To close this gap, subsequent efforts have explored few-shot in-context
learning~\citep{yang2024faima}, chain-of-thought
prompting~\citep{wei2022chain}, instruction
tuning~\citep{scaria2024instructabsa}, and decomposed pipelines that pair
fine-tuned extractors with an LLM-as-a-judge~\citep{bodke2025pastel}; a
parallel line of work investigates multi-agent debate for related
information extraction tasks~\citep{du2023improving,lu2025crossagentie}. However, none of these methods explicitly
mitigate the systematic single-pass failure modes shown in
Figure~\ref{fig:task} (\textit{bottom}).

To address the challenges of previous methods, we propose \textbf{MASTE}, a training-free
\textbf{M}ulti-\textbf{A}gent pipeline for zero-shot
\textbf{A}spect \textbf{S}entiment \textbf{T}riplet \textbf{E}xtraction.
MASTE decomposes
ASTE into four \emph{sequentially-conditioned} agents---Aspect, Opinion,
Sentiment, and Consistency---that share one backbone LLM, with each
downstream agent receiving the input sentence together with the structured
outputs of all upstream agents.

We conduct experiments on four commonly used ASTE benchmarks and compare
MASTE against a comprehensive set of previous approaches.
Our results show that MASTE achieves state-of-the-art performance among
LLM-based methods on all four datasets.
We further perform cross-backbone and ablation studies to verify MASTE's generality and assess each component's contribution.
Overall, our contributions are summarized as follows:
\begin{itemize}
  \item \textbf{A training-free multi-agent pipeline for ASTE.} To the best of our knowledge, MASTE is the first training-free multi-agent pipeline that
requires no task-specific fine-tuning and exposes intermediate structures that can be validated and reused by downstream agents.

  \item \textbf{Span-aware agent design for exact-match ASTE.} Motivated by observed systematic failure modes of zero-shot LLM extraction, we design two mechanisms: minimal-span opinion extraction
with one-to-many aspect--opinion mapping and a Consistency Check Agent that performs span grounding, polarity calibration, and duplicate consolidation over the full triplet set.

  \item \textbf{Comprehensive evaluation on ASTE benchmarks.} We conduct extensive experiments on multiple benchmark datasets to validate
the effectiveness of MASTE. Results show that the proposed pipeline consistently
improves LLM-based ASTE, while ablation and cross-backbone analyses
confirm the importance and generality of its stage-wise design.

\end{itemize}

\begin{figure*}[t!]
    \centering
    \includegraphics[width=0.99\linewidth]{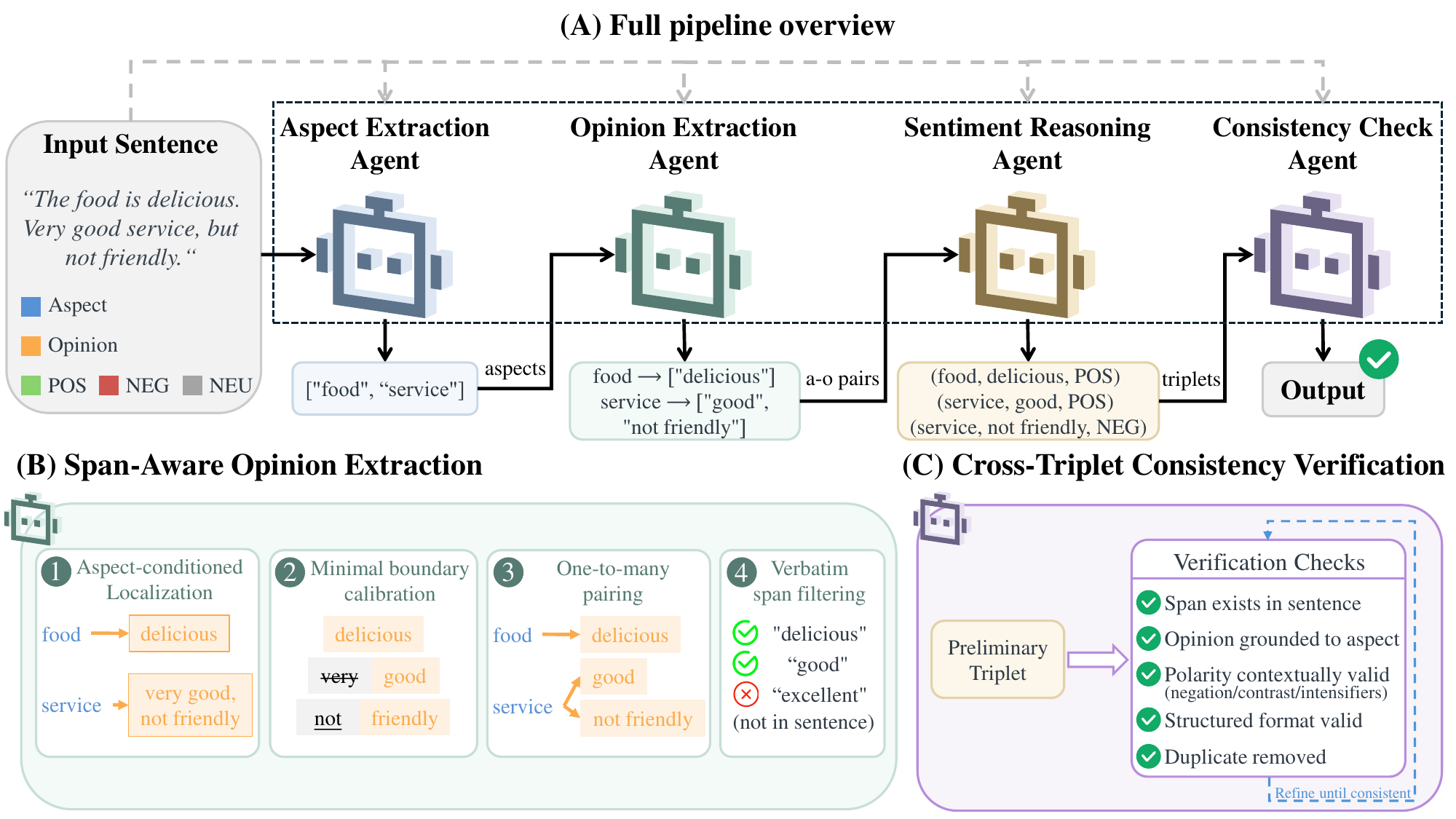}
    \caption{The MASTE pipeline on an example from ASTE-Data-V2.
  \textbf{Solid arrows} carry structured data between stages;
  \textbf{dashed arrows} pass the original sentence as context to each agent.
  Stage~2 enforces \emph{minimal} opinion spans and supports
  one-to-many aspect--opinion mapping.
  Stage~4 validates the full triplet set: hallucination removal, polarity calibration, duplicate consolidation, and output canonicalization.}
    \label{fig:pipeline}
\end{figure*}

\section{Related Work}

\subsection{Aspect Sentiment Triplet Extraction}

Aspect Sentiment Triplet Extraction (ASTE) was introduced by \citet{peng2020knowing} to jointly identify aspect terms, opinion
expressions, and sentiment polarities.
Most prior work relies on supervised learning, including
position-aware tagging~\citep{xu2020position}, sequence-to-sequence generation~\citep{zhang2021towards,yan2021unified}, grid-tagging
schemes~\citep{wu2020grid,chen2022enhanced,sun2024miniconGTS},
transformer-based span modeling~\citep{naglik2024aste}, and multi-domain
joint training~\citep{hou2024trainonce}.
Although effective in-domain, these methods require task-specific triplet annotations and
typically learn dataset-specific span conventions. MASTE instead studies
the training-free setting, where the system must identify exact ASTE
triplets without updating model parameters.

Recent work has begun to explore LLMs for ABSA and ASTE. MiniConGTS~\citep{sun2024miniconGTS} evaluates GPT-3.5-turbo and GPT-4 on ASTE and shows that
pretraining--finetuning remains competitive even in the era of LLMs. Other LLM-based ABSA methods use instruction tuning~\citep{scaria2024instructabsa}, feature-aware
in-context learning~\citep{yang2024faima}, or syntax--opinion--sentiment
reasoning chains~\citep{fan2025aspect}. 
Closest to our setting,
PASTEL~\citep{bodke2025pastel} decomposes ASTE into subtasks and applies an LLM-as-a-judge, but still depends on finetuned extractors before validation. 
In contrast, MASTE performs decomposition and span-aware validation entirely with training-free LLM agents.

\subsection{Multi-Agent LLMs for Information Extraction}

Multi-agent LLM systems were popularized by debate-style methods, where
multiple model instances critique and revise one another's outputs
\citep{du2023improving}. 
This idea has recently been adapted to structured
information extraction: CROSSAGENTIE~\citep{lu2025crossagentie} performs cross-type and cross-task
debate between Named Entity Recognition (NER) and Relation Extraction (RE) agents;
AgentRE~\citep{shi2024agentre} equips an LLM
with memory, retrieval, and reflection modules for low-resource relation
extraction; and Agent-Event-Coder~\citep{wang2025eventcoder} treats zero-shot
event extraction as a multi-agent code-generation and verification
process. MASTE differs from these systems by using sequentially conditioned agents over
typed ASTE intermediates and by targeting span boundary control for
aspect--opinion--sentiment triplets.

\subsection{Decomposition and Verification}

Decomposition and verification are common strategies for improving LLM
reliability. Chain-of-thought prompting~\citep{wei2022chain} elicits intermediate reasoning within a
single generation, while least-to-most prompting~\citep{zhou2023least} decomposes complex problems into ordered subproblems. Verification methods further check generated
solutions using learned verifiers or agentic verification procedures
\citep{cobbe2021training,verifiagent2025}. MASTE adapts these ideas to
fine-grained structured extraction: instead of
verifying a free-form final answer, the Consistency Check Agent operates over the whole predicted
triplet set, removing hallucinated spans, calibrating sentiment labels, and
normalizing opinion boundaries for ASTE's exact-match evaluation.

\section{Method}

\subsection{Task Definition}

Given an input sentence $s = (w_1, w_2, \ldots, w_n)$ of $n$ tokens, the
goal of ASTE is to extract the complete set of triplets: 
\begin{equation}
    \mathcal{T}(s) \;=\; \{(a_i,\,o_i,\,p_i)\}_{i=1}^{m}
\end{equation}
where $a_i$ is an aspect term appearing in $s$,
$o_i$ is the corresponding opinion term, and $p_i \in \{\texttt{POS}, \texttt{NEG}, \texttt{NEU}\}$ denotes the sentiment
polarity.
$m$ is the number of sentiment triplets contained in sentence $s$.
Following standard ASTE evaluation~\citep{xu2020position}, a
prediction is correct only when the aspect term, opinion term, and sentiment polarity
all exactly match a gold triplet.

\subsection{Overall Framework}

Figure~\ref{fig:pipeline} illustrates the overall framework of MASTE. MASTE decomposes ASTE into
four sequentially conditioned agents that share the same frozen LLM backbone $\mathcal{M}$ and operate on typed intermediate outputs.
Each downstream agent receives the original sentence together with the structured outputs of all upstream agents. The following paragraphs describe each agent in detail.

\paragraph{Aspect Extraction Agent.}
The Aspect Extraction Agent $f_{\text{asp}}$ identifies aspects mentioned in the input sentence. Given $s$, it returns a set of aspect terms:
\begin{equation}
  f_{\text{asp}}(s;\,\mathcal{M}) \;\rightarrow\; \mathcal{A}=\{a_1,\ldots,a_k\}
\end{equation}
We constrain the agent to extract only aspect spans that are explicitly
mentioned in $s$, copy them verbatim from the sentence, and avoid
paraphrasing, lemmatization, or boundary expansion. Since all downstream
stages condition on $\mathcal{A}$, errors at this stage can propagate
through the entire pipeline. We therefore apply deterministic span
validation after generation: any returned aspect that does not appear as
a case-insensitive substring of $s$ is discarded before opinion
extraction.

\paragraph{Opinion Extraction Agent.}
Conditioned on the sentence $s$ and the validated aspect set
$\mathcal{A}$, the Opinion Extraction Agent $f_{\text{opn}}$ identifies
opinion spans that are associated with each extracted aspect:
\begin{equation}
  f_{\text{opn}}(s,\,\mathcal{A};\,\mathcal{M})
  \;\rightarrow\;
  \mathcal{P}
  =
  \{(a_i, o_i^{(j)})\}
\end{equation}
Here $o_i^{(j)}$ denotes the $j$-th opinion span associated with aspect
$a_i$, allowing the same aspect to appear in multiple pairs when it is
described by multiple independent opinions. 
We further introduce a span-aware extraction procedure consisting of four steps, which is detailed in \S\ref{sec:method:opinion}.

\paragraph{Sentiment Reasoning Agent.}
The Sentiment Reasoning Agent $f_{\text{sen}}$ takes the sentence $s$ and
the aspect--opinion pair set $\mathcal{P}$, and assigns a sentiment
polarity to each pair:
\begin{equation}
  f_{\text{sen}}(s,\,\mathcal{P};\,\mathcal{M})
  \;\rightarrow\;
  \mathcal{T}^{(3)}
  =
  \{(a_i, o_i^{(j)}, p_i^{(j)})\}
\end{equation}
where $p_i^{(j)} \in \{\texttt{POS},\texttt{NEG},\texttt{NEU}\}$ denotes
the polarity of opinion $o_i^{(j)}$ toward aspect $a_i$. By conditioning
on the full sentence rather than the opinion span alone, the agent can
account for contextual polarity cues such as negation, intensification,
and contrast. The resulting set $\mathcal{T}^{(3)}$ serves as the
preliminary triplet set for the Consistency Check Agent.

\paragraph{Consistency Check Agent.}
The Consistency Check Agent $f_{\text{con}}$ refines the
preliminary triplet set $\mathcal{T}^{(3)}$ in the context of the
original sentence $s$, yielding the final calibrated triplet set:
\begin{equation}
  f_{\text{con}}(s,\,\mathcal{T}^{(3)};\,\mathcal{M})
  \;\rightarrow\;
  \mathcal{T}.
\end{equation}
Unlike single-pass prompting or per-triplet verification, $f_{\text{con}}$
calibrates the output at the triplet-set level, allowing residual errors to be corrected. We detail this procedure in
\S\ref{sec:method:consistency}.

\subsection{Span-Aware Opinion Extraction}
\label{sec:method:opinion}
To address common LLM failure modes in ASTE, especially span bloat and
triplet merging (Figure~\ref{fig:task},~\textit{bottom}), we design
$f_{\text{opn}}$ as a span-aware extraction stage with the following
procedure:

\begin{enumerate}[noitemsep, topsep=2pt, leftmargin=1.6em]
  \item \textbf{Aspect-conditioned opinion localization.}
  For each aspect $a_i \in \mathcal{A}$, identify all opinion expressions
  in $s$ that describe $a_i$.

  \item \textbf{Minimal boundary calibration.}
    Select the shortest contiguous span that preserves the core opinion.
    For example, degree-only modifiers are removed
    (\emph{``very clean''}$\rightarrow$\emph{``clean''}), whereas
    polarity-bearing expressions are preserved
    (\emph{``not bad''}$\rightarrow$\emph{``not bad''}).

  \item \textbf{One-to-many opinion pairing.}
  For each independent opinion associated with $a_i$, construct a separate
  pair $(a_i, o_i^{(j)})$, allowing the same aspect to appear in multiple
  pairs.

  \item \textbf{Verbatim span filtering.}
  Drop pairs whose aspect is not in $\mathcal{A}$, whose opinion span
  does not appear verbatim in $s$, or that duplicate an earlier pair.
\end{enumerate}

\begin{table*}[t!]
\centering
\small
\setlength{\tabcolsep}{4.5pt}
\renewcommand{\arraystretch}{1.08}
\resizebox{\textwidth}{!}{
\begin{tabular}{ccccccccccccc}
\toprule
\multirow{2}{*}[-1.0ex]{\textbf{Method}}
& \multicolumn{3}{c}{\textbf{14Res}}
& \multicolumn{3}{c}{\textbf{14Lap}}
& \multicolumn{3}{c}{\textbf{15Res}}
& \multicolumn{3}{c}{\textbf{16Res}} \\
\cmidrule(lr){2-4}
\cmidrule(lr){5-7}
\cmidrule(lr){8-10}
\cmidrule(lr){11-13}
& P & R & F1
& P & R & F1
& P & R & F1
& P & R & F1 \\
\midrule

\multicolumn{1}{c}{\textbf{Pipeline}} 
& \multicolumn{12}{c}{} \\

CMLA+ \cite{wang2017coupled}
& 39.18 & 47.13 & 42.79
& 30.09 & 36.92 & 33.16
& 34.56 & 39.84 & 37.01
& 41.34 & 42.10 & 41.72 \\

RINANTE+ \cite{dai2019neural}
& 31.42 & 39.38 & 34.95
& 21.71 & 18.66 & 20.07
& 29.88 & 30.06 & 29.97
& 25.68 & 22.30 & 23.87 \\

Two-stage \cite{peng2020knowing}
& 43.24 & 63.66 & 51.46
& 37.38 & \underline{50.38} & 42.87
& \underline{48.07} & 57.51 & \underline{52.32}
& 46.96 & 64.24 & 54.21 \\

Li-unified-R+PD \cite{peng2020knowing}
& 40.56 & 44.28 & 42.34
& 41.04 & \textbf{67.35} & \underline{51.00}
& 44.72 & 51.39 & 47.82
& 37.33 & 54.51 & 44.31 \\

\midrule
\multicolumn{1}{c}{\textbf{LLM-based}} 
& \multicolumn{12}{c}{} \\
GPT-4o zero-shot
& 32.99 & 38.13 & 35.37
& 17.81 & 22.55 & 19.90
& 27.85 & 37.73 & 32.05
& 32.17 & 43.00 & 36.80 \\

GPT-4o few-shot
& \underline{54.11} & \underline{66.20} & \underline{59.55}
& \underline{38.23} & 48.61 & 42.80
& 45.57 & \underline{60.41} & 51.95
& \underline{52.90} & \underline{71.01} & \underline{60.63} \\

GPT-4o CoT
& 41.21 & 53.32 & 46.49
& 26.98 & 37.71 & 31.46
& 33.07 & 50.93 & 40.10
& 39.14 & 58.17 & 46.79 \\

GPT-4o CoT+few-shot
& 46.81 & 59.86 & 52.54
& 29.71 & 40.85 & 34.40
& 35.08 & 53.81 & 42.47
& 41.53 & 61.09 & 49.45 \\
\midrule

\multicolumn{1}{c}{\textbf{Ours}} 
& \multicolumn{12}{c}{} \\

GPT-4o MASTE
& \textbf{74.64} & \textbf{70.12} & \textbf{72.31}
& \textbf{64.78} & 49.24 & \textbf{55.95}
& \textbf{68.95} & \textbf{66.04} & \textbf{67.46}
& \textbf{74.48} & \textbf{72.35} & \textbf{73.40} \\

\bottomrule
\end{tabular}}
\caption{
Experimental results on ASTE-Data-V2~\citep{xu2020position}. 
Precision, Recall, and F1 score are reported for each dataset.
The best results are highlighted in bold, and the second-best results are underlined.}
\label{tab:main_results}
\end{table*}

\subsection{Cross-Triplet Consistency Verification}
\label{sec:method:consistency}

The first three agents yield a preliminary triplet set
$\mathcal{T}^{(3)}$, but residual errors such as unsupported spans, polarity mismatches, duplicate predictions, and invalid output formats may remain. We therefore use $f_{\text{con}}$ as a triplet-set-level verifier,
which performs four consistency checks conditioned on $s$ and $\mathcal{T}^{(3)}$:

\begin{enumerate}[noitemsep, topsep=2pt, leftmargin=1.6em]
  \item \textbf{Span grounding.}
  Remove triplets whose aspect or opinion span does not appear verbatim in
  $s$.
  \item \textbf{Polarity calibration.}
    Revise sentiment labels that conflict with contextual cues such as negation or contrast.
    \item \textbf{Duplicate consolidation.}
      Merge duplicate or semantically redundant triplets.
    \item \textbf{Output canonicalization.}
    Convert the verified result into a canonical triplet format used for
    evaluation, with normalized sentiment labels and valid structured output.
\end{enumerate}

\section{Experiments}
\subsection{Datasets and Metrics}
We conduct experiments on four public datasets, \textsc{Lap14},
\textsc{Res14}, \textsc{Res15}, and \textsc{Res16}, which are derived from
the sentiment evaluation benchmarks SemEval 2014~\citep{pontiki2014semeval},
SemEval 2015~\citep{pontiki2015semeval}, and
SemEval 2016~\citep{pontiki2016semeval}, respectively. There are two commonly used versions of these datasets in prior ASTE studies. Following \citet{xu2020position}, we
conduct all experiments on ASTE-Data-V2, with dataset statistics summarized in Appendix~\ref{app:implementation_details}.

We evaluate all models using the widely accepted (\texttt{Precision}, \texttt{Recall}, \texttt{F1}) metrics.

\subsection{Baselines}
We compare MASTE with two groups of baselines: (1) pipeline-based methods, which decompose ASTE into intermediate subtasks or rely on extracted intermediate
structures. These methods provide relevant
comparisons for evaluating the stage-wise design of MASTE. (2) LLM-based methods with different prompting strategies. Notably, MASTE uses the same backbone (GPT-4o) as the LLM-based baselines,
but does not require any in-context demonstrations.

\subsection{Main Results}
Table~\ref{tab:main_results} summarizes the results comparing MASTE (GPT-4o) with all
pipeline-based and LLM-based baselines on
ASTE-Data-V2. The best results are highlighted in bold, and the second-best results are underlined. Overall, MASTE achieves the best F1 score among the compared methods on all four datasets.

\paragraph{Comparison with pipeline-based methods.}
Compared with pipeline-based baselines, MASTE achieves the highest F1 score on all four datasets, but the margin varies across domains. It improves over the strongest pipeline baseline by 20.85 F1 points on \textsc{14Res}, 4.95 points on \textsc{14Lap}, 15.14 points on \textsc{15Res}, and 19.19 points on \textsc{16Res}. The gain is therefore substantial on the restaurant datasets, while more moderate on \textsc{14Lap}, where Li-unified-R+PD obtains much higher recall. This pattern indicates that MASTE is not simply more recall-oriented than prior pipeline systems; rather, its advantage mainly comes from producing more precise exact-match triplets. In particular, MASTE achieves the highest precision on every dataset, suggesting that span grounding, aspect-conditioned opinion extraction, and triplet-level consistency checking help reduce unsupported spans and incorrect aspect--opinion pairings. On \textsc{14Lap}, the lower recall also reveals a remaining limitation: MASTE may miss valid triplets in domains where pipeline systems recover broader candidate sets.

\paragraph{Comparison with LLM-based methods.}
MASTE also consistently outperforms all GPT-4o prompting baselines. Compared
with zero-shot prompting, MASTE improves F1 by 36.94, 36.05, 35.41, and 36.60
points on \textsc{14Res}, \textsc{14Lap}, \textsc{15Res}, and \textsc{16Res},
respectively. Compared with few-shot prompting, MASTE still achieves F1 gains of 12.76, 13.15, 15.51, and 12.77 points on the four datasets, without using in-context demonstrations. This
suggests that the gains come from the proposed multi-agent decomposition
rather than additional demonstration examples.

\paragraph{Precision-oriented improvements.}
A consistent pattern in Table~\ref{tab:main_results} is that MASTE achieves the
highest precision on all four datasets. This is important for ASTE because
exact-match evaluation penalizes hallucinated terms, over-extended opinion
spans, and incorrect aspect--opinion pairings. The large precision gains suggest
that the Aspect Extraction Agent and Consistency Check Agent effectively filter
unsupported spans and reduce false positives. Although MASTE does not always
obtain the highest recall, especially on \textsc{14Lap}, it maintains competitive
recall while producing more reliable exact-match triplets.

\paragraph{Effect of decomposition.}
The results also show that chain-of-thought prompting alone is insufficient for
ASTE. Although CoT encourages step-by-step reasoning, it still performs the
entire extraction process within a single generation. In contrast, MASTE assigns
aspect extraction, opinion extraction, sentiment reasoning, and consistency
verification to separate agents with structured intermediate outputs. This
design reduces the burden on a single decoding pass and better matches the
compositional nature of ASTE.

\section{Analysis}

\begin{table}[t]
\centering
\small
\setlength{\tabcolsep}{3.5pt}
\renewcommand{\arraystretch}{1.08}
\resizebox{\columnwidth}{!}{
\begin{tabular}{lccccc}
\toprule
\textbf{Ablation Setting}
& \textbf{14Res} & \textbf{14Lap}
& \textbf{15Res} & \textbf{16Res}
& \textbf{Avg.} \\
\midrule
MASTE (full)
& \textbf{72.31} & \textbf{55.95} & \textbf{67.46} & \textbf{73.40} & \textbf{67.28} \\
w/o Consistency
& 68.04 & 54.37 & 59.86 & 67.59 & 62.47 \\
w/o Opinion
& 60.90 & 51.69 & 59.72 & 64.48 & 59.20 \\
w/o Sentiment
& 62.84 & 50.12 & 57.45 & 65.94 & 59.09 \\
w/o Opinion + Consistency
& 40.35 & 36.65 & 41.84 & 44.18 & 40.76 \\
\bottomrule
\end{tabular}}
\caption{Ablation study on F1 scores with GPT-4o where w/o denotes removal of the corresponding agent from MASTE. Full results are provided in Appendix~\ref{app:full_ablation}.}
\label{tab:ablation}
\end{table}

\begin{table}[t!]
\centering
\small
\setlength{\tabcolsep}{3.2pt}
\renewcommand{\arraystretch}{1.08}
\resizebox{\columnwidth}{!}{
\begin{tabular}{lcccccc}
    \toprule
    \textbf{Backbone} & \textbf{Method}
    & \textbf{14Res} & \textbf{14Lap}
    & \textbf{15Res} & \textbf{16Res}
    & \textbf{Avg.} \\
    \midrule
    \multirow{3}{*}{GPT-3.5-turbo}
    & Zero-shot  &46.18 &31.88 &39.21 &44.34 & 40.40 
    \\
    & MASTE     & 57.41 & 36.45 & 50.93 & 56.19 & 50.25  \\
    & $\Delta$  & 11.23 & 4.57 & 11.72 & 11.85 & 
    \\
    
    \midrule
    \multirow{3}{*}{Claude Sonnet 4.6}
    & Zero-shot & 61.06 & 38.59 & 54.09 & 63.01 & 54.19 \\
    & MASTE     & 66.41 & 48.23 & 56.70 & 65.83 & 59.29\\
    & $\Delta$  & 5.35 & 9.64 & 2.61 & 2.82 & 
    \\

    \midrule
    \multirow{3}{*}{Gemini-3-flash}
    & Zero-shot & 54.36 & 35.85 & 46.89 & 56.97 & 48.52 \\
    & MASTE     & 67.48 & 51.80 & 62.98 & 67.74 & 62.50\\
    & $\Delta$  & 13.12 & 15.95 & 16.09 & 10.77 & 
    \\

 \midrule
    \multirow{3}{*}{Deepseek-V3.2}
    & Zero-shot & 47.55 & 31.22 & 41.86 & 50.93 & 42.89 \\
    & MASTE     & 66.38 & 49.32 & 57.25 & 64.39 & 59.33\\
    & $\Delta$  & 18.83 & 18.10 & 15.39 & 13.46 & 
    \\

     \midrule
    \multirow{3}{*}{Seed 2.0 pro}
    & Zero-shot & 39.02 & 27.31 & 32.84 & 36.96 & 34.03 \\
    & MASTE     & 67.60 & 49.16 & 62.71 & 65.32 & 61.20\\
    & $\Delta$  & 28.58 & 21.85 & 29.87 & 28.36 & 
    \\
    \bottomrule
\end{tabular}}

\caption{Cross-backbone analysis on F1 scores. $\Delta$ denotes the absolute F1 gain of MASTE over Zero-shot under the same backbone. Full results are provided in Appendix~\ref{app:full_crossbackbone}.}
\label{tab:crossbackbone}
\end{table}

\subsection{Ablation Study}

We conduct ablation experiments to quantify the contribution of each agent in
MASTE. Table~\ref{tab:ablation} reports F1 scores on the four datasets using
GPT-4o as the backbone. The full model uses all four agents. The ablated
variants remove the Consistency Check Agent, the Opinion Extraction Agent, the
Sentiment Reasoning Agent, or both the Opinion Extraction and Consistency Check Agents.

The full model achieves the best F1 on all datasets, with an average F1 of
67.28. Removing the Consistency Check Agent reduces the average F1 from 67.28 to 62.47. The drop is observed on all datasets, indicating that final
triplet-level verification is consistently useful. This stage helps remove
unsupported spans, duplicated triplets, and invalid aspect--opinion--sentiment
combinations before evaluation.

Removing the Opinion Extraction Agent causes a larger average drop. This result confirms the importance of aspect-conditioned opinion localization. Without this stage, the model is more likely to miss
aspect-specific opinion expressions or generate opinion spans with inaccurate
boundaries. Removing the Sentiment Reasoning Agent also leads to a clear degradation, suggesting that explicit sentiment reasoning remains significant even when the aspect and opinion spans
are already identified.

The largest degradation appears when the Opinion Extraction and Consistency Check Agents are removed together. In this setting, the average F1 drops sharply from
67.28 to 40.76. This
result shows that local
aspect-conditioned opinion extraction and global consistency verification are complementary. 
The former constructs higher-quality candidate pairs, while the
latter filters remaining invalid triplets. Overall, the ablation results confirm that the four agents contribute jointly to ASTE.

\newcommand{\golderr}[1]{{\color{green!45!black}#1}}
\newcommand{\prederr}[1]{{\color{red!70!black}#1}}

\begin{table*}[t]
\centering
\small
\setlength{\tabcolsep}{3pt}
\renewcommand{\arraystretch}{1.08}
\begin{tabular}{L{0.27\textwidth}C{0.27\textwidth}C{0.27\textwidth}C{0.16\textwidth}}
\toprule
\textbf{Sentence} & \textbf{Gold Triplet(s)} & \textbf{MASTE Prediction} & \textbf{Error Type} \\
\midrule
\lcell{\emph{It is fast and easy to use.}}
&
\ccell{\golderr{(use, easy, POS)}}
&
\ccell{\prederr{$\emptyset$}}
&
\ccell{Missing prediction} \\
\midrule
\lcell{\emph{The baterry is very longer.}}
&
\ccell{(\golderr{baterry}, longer, POS)}
&
\ccell{(\prederr{battery}, longer, POS)}
&
\ccell{Span normalization} \\
\midrule
\lcell{\emph{I am pleased with the fast log on, speedy WiFi connection and the long
battery life.}}
&
\ccell{
(\golderr{log on}, \golderr{pleased}, POS);\\
(\golderr{WiFi connection}, \golderr{pleased}, POS);\\
(\golderr{battery life}, \golderr{pleased}, POS)}
&
\ccell{
(log on, fast, POS);\\
(WiFi connection, speedy, POS);\\
(battery life, long, POS);\\
\prederr{$\emptyset$ for shared opinion links}}
&
\errcell{Shared-opinion\\omission}\\ 
\midrule
\lcell{\emph{No green beans, no egg, no anchovy dressing, no nicoise olives, no red onion.}}
&
\ccell{
(\golderr{green beans}, No, \golderr{NEU});\\
(\golderr{egg}, no, \golderr{NEU});\\
...}
&
\ccell{
(\prederr{green beans}, no, \prederr{NEG});\\
(\prederr{egg}, no, \prederr{NEG});\\
...}
&
\errcell{Polarity convention} \\
\midrule

\lcell{\emph{The food is all-around good, with the rolls usually excellent and the
sushi/sashimi not quite on the same level.}}
&
\errcell{
(food, good, POS);\\
(rolls, excellent, POS)}
&
\ccell{(food, good, POS);\\
(rolls, excellent, POS);\\
\prederr{(sushi/sashimi, not quite}\\ 
\prederr{on the same level, NEG)}}
&
\errcell{Annotation-\\sensitive FP} \\
\bottomrule
\end{tabular}
\caption{Representative errors made by the full MASTE pipeline. The analysis
focuses on residual failure modes. Green marks
gold-only or gold-correct elements; red marks missing, mismatched, or extra
MASTE predictions.}
\label{tab:error_analysis}
\end{table*}

\subsection{Cross-Backbone Analysis}
We further evaluate whether MASTE generalizes across different LLM backbones.
Table~\ref{tab:crossbackbone} compares direct zero-shot prompting with MASTE (which also requires no in-context demonstrations)
under multiple backbone models.

Across all backbone settings, MASTE consistently outperforms direct
zero-shot prompting on all four datasets. 
The gains are observed across representative models from different model families, indicating that the effectiveness of MASTE
does not depend on a single proprietary model and can generalize across diverse backbones. 
These results further
support our central hypothesis that decomposing ASTE into specialized
subtasks provides a more reliable inference process than single-pass triplet generation.

Overall, the cross-backbone results support the generality of MASTE as a
training-free and backbone-agnostic framework for zero-shot ASTE.

\subsection{Error Analysis}
The ablation study quantifies the contribution of each agent to the final
performance. We further inspect the residual errors made by the full MASTE
pipeline. Table~\ref{tab:error_analysis} presents representative failure cases.
Following the order of the examples, these errors mainly arise from
(i) missed event-like aspect spans, (ii) surface-form mismatches under
exact-match evaluation, (iii) incomplete recovery of shared opinion links,
(iv) polarity disagreement with dataset-specific annotation conventions, and
(v) annotation-sensitive extractions that are textually plausible but absent
from the gold labels.

The first two cases are span-level failures. In the first case, the gold aspect
\emph{use} is predicate-like rather than a concrete noun phrase. Once the
Aspect Extraction Agent misses this span, all downstream agents receive no
valid target and MASTE returns an empty set. In the second case, MASTE
normalizes the misspelled gold aspect \emph{baterry} into \emph{battery}. The
prediction is semantically close, but exact-match evaluation requires the
surface span to be copied exactly, so the triplet is counted as incorrect.

The third case shows a relation-coverage error. MASTE extracts local attribute
opinions such as \emph{fast}, \emph{speedy}, and \emph{long}, but misses the
shared opinion \emph{pleased}, which applies to multiple aspects. This suggests
that one-to-many aspect--opinion recovery remains difficult when a sentence
contains both local modifiers and a global evaluative predicate.

The last two cases are annotation-sensitive. For absence expressions such as
\emph{no}, MASTE tends to infer negative sentiment, while ASTE-Data-V2 may
annotate the corresponding triplets as neutral. Conversely, MASTE may extract a
textually supported negative comparison absent from the gold set.
These cases suggest that future work should focus not only on aspect
recall, but also on annotation-aware polarity calibration and uncertainty
handling for convention-dependent triplets.

\section{Conclusion}
In this paper, we presented MASTE, a training-free multi-agent framework for
zero-shot Aspect Sentiment Triplet Extraction. Rather than requiring a single
LLM generation to determine aspect spans, opinion spans, sentiment labels, and
their pairings at once, MASTE decomposes ASTE into four sequentially
conditioned agents for aspect extraction, aspect-conditioned opinion
extraction, sentiment reasoning, and triplet-level consistency verification.
Experiments on four benchmark datasets from ASTE-Data-V2 show that MASTE
consistently improves over existing pipeline-based and LLM-based
methods. Ablation and cross-backbone analyses further confirm the contribution of
each stage and the generality of the framework across diverse LLM families.
These findings suggest that explicit intermediate
structures and verification are effective tools for adapting
LLMs to fine-grained structured sentiment extraction without task-specific
training.

\section{Limitations}

MASTE's performance is 
inherently bounded by the reasoning and span-extraction capabilities 
of the underlying backbone LLM.
In particular, errors made in the Aspect Extraction Agent can propagate
to all downstream stages, since opinion extraction and sentiment reasoning
both condition on the predicted aspect set. Therefore, the aspect recall remains a major bottleneck.


The multi-agent design introduces additional inference cost relative to
single-pass generation. MASTE requires four LLM calls per sentence, which
increases both latency and monetary cost in large-scale or real-time
applications. Future work may explore model-level routing mechanisms that
assign simpler inputs or subtasks to smaller models while reserving stronger
LLMs for ambiguous and complex cases. Such a method could improve efficiency while preserving the benefits of staged
decomposition.

Finally, our experiments are conducted on English review-domain ASTE
benchmarks. Although the cross-backbone results support the robustness of the
framework across LLM families, further validation is needed for
multilingual settings, longer documents, noisier user-generated content, and
domains whose annotation conventions differ from ASTE-Data-V2. We encourage future work to evaluate MASTE in these broader conditions.

\section{Ethical considerations}

This work is designed for scientific research on aspect-level sentiment
analysis. Our experiments are conducted on public benchmark datasets and do
not involve new user data collection or human subject experiments. We believe MASTE
can contribute positively to the ABSA research community by reducing the
dependence on task-specific training data, improving the transparency of
intermediate extraction decisions, and providing reusable prompts and code for
future research.

At the same time, MASTE inherits potential risks from the underlying LLMs. Its outputs may reflect
model bias or unsupported sentiment
inferences, and therefore should not be treated as ground-truth judgments in
high-stakes decision making. When applying the framework to proprietary,
private, or sensitive text, practitioners should ensure that data handling
complies with the privacy policies and legal requirements of their deployment
context. We will release code and prompts to facilitate reproducibility and to
make each extraction stage easier to audit.

\bibliography{arxiv}

\newpage
\appendix

\section{Appendix}
\label{sec:appendix}

\subsection{Implementation Details}
\label{app:implementation_details}

For closed-source LLM backbones, we use the official provider APIs, including
OpenAI for GPT models, Anthropic for Claude, Google for Gemini, etc. Unless
otherwise specified, all LLM calls use temperature $0$ and all intermediate
outputs are parsed as JSON objects before being passed to downstream agents or
the evaluator.

We conduct experiments on ASTE-Data-V2~\citep{xu2020position}, which contains
four standard ASTE domains: \textsc{14Lap} (Lap14), \textsc{14Res} (Res14),
\textsc{15Res}, and \textsc{16Res}. These domains are derived from SemEval
2014~\citep{pontiki2014semeval}, SemEval 2015~\citep{pontiki2015semeval}, and
SemEval 2016~\citep{pontiki2016semeval}. Table~\ref{tab:dataset_statistics}
summarizes the datasets.

All systems are evaluated with micro-averaged exact-match precision, recall,
and F1 over triplet sets. Before matching, aspect and opinion strings are
case-normalized and whitespace-normalized, and sentiment labels are mapped to a
canonical uppercase form. A predicted triplet is counted as correct only when
its normalized aspect span, normalized opinion span, and sentiment label all
match a gold triplet in the same sentence. Duplicate predictions collapse under
set matching, so repeated identical triplets do not increase the number of
correct predictions.

\begin{table}[h]
\centering
\scriptsize
\setlength{\tabcolsep}{3.5pt}
\renewcommand{\arraystretch}{1.05}
\begin{tabular}{llrrrrrrrrrr}
\toprule
\textbf{Dataset} & \textbf{Split} & \textbf{\#S} & \textbf{\#T} & \textbf{\#A} & \textbf{\#O} & \textbf{\#+} & \textbf{\#0} & \textbf{\#-} & \textbf{\#SW} & \textbf{\#MW} \\
\midrule
\multirow{3}{*}{\textsc{14Res}}
& Train & 1266 & 2338 & 986 & 844 & 1692 & 166 & 480 & 1586 & 752 \\
& Dev   & 310  & 577  & 396 & 307 & 404  & 54  & 119 & 388  & 189 \\
& Test  & 492  & 994  & 579 & 437 & 773  & 66  & 155 & 657  & 337 \\
\midrule
\multirow{3}{*}{\textsc{14Lap}}
& Train & 906 & 1460 & 733 & 695 & 817 & 126 & 517 & 824 & 636 \\
& Dev   & 219 & 346  & 268 & 237 & 169 & 36  & 141 & 190 & 156 \\
& Test  & 328 & 543  & 400 & 329 & 364 & 63  & 116 & 291 & 252 \\
\midrule
\multirow{3}{*}{\textsc{15Res}}
& Train & 605 & 1013 & 582 & 462 & 783 & 25 & 205 & 678 & 335 \\
& Dev   & 148 & 249  & 191 & 183 & 185 & 11 & 53  & 165 & 84  \\
& Test  & 322 & 485  & 347 & 310 & 317 & 25 & 143 & 297 & 188 \\
\midrule
\multirow{3}{*}{\textsc{16Res}}
& Train & 857 & 1394 & 759 & 623 & 1015 & 50 & 329 & 918 & 476 \\
& Dev   & 210 & 339  & 251 & 221 & 252  & 11 & 76  & 216 & 123 \\
& Test  & 326 & 514  & 338 & 282 & 407  & 29 & 78  & 344 & 170 \\
\bottomrule
\end{tabular}
\caption{Dataset statistics of ASTE-Data-V2. \#S, \#T, \#A, and \#O denote
the numbers of sentences, triplets, aspects, and opinions, respectively. \#+,
\#0, and \#- denote positive, neutral, and negative triplets. \#SW and \#MW
denote single-word and multi-word triplets, respectively.}
\label{tab:dataset_statistics}
\end{table}

\subsection{Prompt Templates}
\label{app:prompts}

We report the prompt templates used by both MASTE and the LLM-based
baselines. 
The MASTE prompts follow a staged agent design: each agent receives
only the information required for its subtask, returns a structured JSON
object, and passes validated intermediate outputs to downstream agents. 
For the LLM-based baselines, we
follow the zero-shot, few-shot, CoT, and CoT+few-shot settings used in~\citep{sun2024miniconGTS}.
To ensure that all methods can be evaluated
under the same exact-match protocol, we require the baseline prompts to produce
the same JSON triplet format as MASTE.


\subsubsection{MASTE Prompts}
The variables
\texttt{\{sentence\}}, \texttt{\{aspects\}}, \texttt{\{pairs\}}, and
\texttt{\{triplets\}} are filled at inference time.

\begin{tcolorbox}[
  title=\textbf{Aspect Extraction Agent},
  colback=gray!3,
  colframe=black,
  colbacktitle=black,
  coltitle=white,
  boxrule=0.8pt,
  arc=2pt,
  left=8pt,
  right=8pt,
  top=8pt,
  bottom=8pt,
  fonttitle=\bfseries,
  enhanced,
  breakable
]

\begin{lstlisting}[
  basicstyle=\ttfamily\small,
  columns=fullflexible,
  breaklines=true,
  breakatwhitespace=false,
  breakautoindent=false,
  breakindent=0pt,
  keepspaces=true,
  showstringspaces=false,
  frame=none
]
System:
You are an expert in aspect-based sentiment analysis. Your task is to identify ASPECT TERMS in a sentence.

Rules:
- Extract only aspects that are explicitly present in the sentence, do not infer or paraphrase.
- Copy the aspect span exactly as it appears in the sentence, including any misspellings.
- Use the minimal span that identifies the evaluated target.
- An aspect can be a single word or a multi-word phrase.
- Return a JSON list of strings only. If no aspects are found, return [].

User:
Sentence: {sentence}
Identify all aspect terms. Return a JSON list of strings. If no aspects are found, return [].
\end{lstlisting}
\end{tcolorbox}

\begin{tcolorbox}[
  title=\textbf{Opinion Extraction Agent},
  colback=gray!3,
  colframe=black,
  colbacktitle=black,
  coltitle=white,
  boxrule=0.8pt,
  arc=2pt,
  left=8pt,
  right=8pt,
  top=8pt,
  bottom=8pt,
  fonttitle=\bfseries,
  enhanced,
  breakable
]

\begin{lstlisting}[
  basicstyle=\ttfamily\small,
  columns=fullflexible,
  breaklines=true,
  breakatwhitespace=false,
  breakautoindent=false,
  breakindent=0pt,
  keepspaces=true,
  showstringspaces=false,
  frame=none
]
System:
You are an expert in aspect-based sentiment analysis. Your task is to identify OPINION EXPRESSIONS for given aspect terms in a sentence.

An opinion expression is the SHORTEST word or phrase copied VERBATIM from the sentence
that captures the reviewer's evaluation of a specific aspect. Copy the span exactly as it appears in the text, even if it contains misspellings or typos.

Rules:
- Extract the shortest opinion phrase copied verbatim from the sentence.
- Do not include degree adverbs unless they change meaning.
- If one aspect has multiple independent opinions, output one entry
  per opinion rather than merging them.
- Omit aspects for which no opinion is found.
- Return a JSON list: [{"aspect": ..., "opinion": ...}].

User:
Sentence: {sentence}
Aspects: {aspects}
For each aspect find its opinion expression(s). Use the minimal span - no degree adverbs unless they change sentiment polarity. If one aspect has multiple opinions, output one entry per opinion. Omit aspects with no opinion. 
Return JSON list: 
[{"aspect": ..., "opinion": ...}].
\end{lstlisting}
\end{tcolorbox}

\begin{tcolorbox}[
  title=\textbf{Sentiment Reasoning Agent},
  colback=gray!3,
  colframe=black,
  colbacktitle=black,
  coltitle=white,
  boxrule=0.8pt,
  arc=2pt,
  left=8pt,
  right=8pt,
  top=8pt,
  bottom=8pt,
  fonttitle=\bfseries,
  enhanced,
  breakable
]

\begin{lstlisting}[
  basicstyle=\ttfamily\small,
  columns=fullflexible,
  breaklines=true,
  breakatwhitespace=false,
  breakautoindent=false,
  breakindent=0pt,
  keepspaces=true,
  showstringspaces=false,
  frame=none
]
System:
You are an expert sentiment analyst. Your task is to determine the SENTIMENT POLARITY for aspect-opinion pairs.

Sentiment labels:
- POS: positive sentiment.
- NEG: negative sentiment.
- NEU: neutral sentiment.

Rules:
- Consider the context of the full sentence, not just the opinion word in isolation.
- Be careful with negation.
- Consider intensifiers and modifiers.

User:
Sentence: {sentence}
Aspect-Opinion pairs: {pairs}
Classify each as POS, NEG, or NEU.
Return JSON list:
[{"aspect": ..., "opinion": ..., "sentiment": ...}].
\end{lstlisting}
\end{tcolorbox}

\begin{tcolorbox}[
  title=\textbf{Consistency Check Agent},
  colback=gray!3,
  colframe=black,
  colbacktitle=black,
  coltitle=white,
  boxrule=0.8pt,
  arc=2pt,
  left=8pt,
  right=8pt,
  top=8pt,
  bottom=8pt,
  fonttitle=\bfseries,
  enhanced,
  breakable
]

\begin{lstlisting}[
  basicstyle=\ttfamily\small,
  columns=fullflexible,
  breaklines=true,
  breakatwhitespace=false,
  breakautoindent=false,
  breakindent=0pt,
  keepspaces=true,
  showstringspaces=false,
  frame=none
]
System:
You are a quality-control expert for aspect-based sentiment analysis. Review a set of extracted (aspect, opinion, sentiment) triplets.

Checks:
1. HALLUCINATION: remove triplets whose aspect or opinion does not appear verbatim in the sentence.
2. SPAN TRIM: trim unnecessary degree adverbs unless they change polarity.
3. SENTIMENT ERROR: fix polarity labels that conflict with context.
4. MISSING: add clearly expressed aspect-opinion pairs that are absent.
5. DUPLICATION: collapse duplicate or redundant triplets.

Rules:
- Default to keeping triplets.
- Do not add speculative triplets unsupported by the sentence.

User:
Sentence: {sentence}
Proposed triplets: {triplets}
Review for hallucinations, sentiment errors, duplicates, and missing pairs. Return only valid, corrected triplets as a JSON list:
[{"aspect": ..., "opinion": ..., "sentiment": ...}].
\end{lstlisting}
\end{tcolorbox}

\subsubsection{Baseline Prompts}
See Table~\ref{tab:baseline_prompts}.

\subsection{Full Ablation Results}
\label{app:full_ablation}
See Table~\ref{tab:appendix_ablation_full}.

\subsection{Full Cross-Backbone Results}
\label{app:full_crossbackbone}
See Table~\ref{tab:appendix_crossbackbone_full}.

\clearpage

\begin{table*}[t]
\centering
\scriptsize
\setlength{\tabcolsep}{4pt}
\renewcommand{\arraystretch}{1.12}
\begin{tabular}{
  >{\centering\arraybackslash}m{0.16\textwidth}
  >{\raggedright\arraybackslash}m{0.78\textwidth}
}
\toprule
\textbf{Baseline} & \textbf{Prompt Template} \\
\midrule
Zero-shot
&
\textbf{Instruction:} Suppose you are an expert in aspect-based sentiment
analysis. Given the input sentence, extract all
\texttt{(aspect, opinion, sentiment)} triplets. The sentiment label must be
\texttt{POS}, \texttt{NEG}, or \texttt{NEU}. Aspect and opinion strings must be
copied verbatim from the sentence. Return only a JSON list in the format
\texttt{[\{"aspect": ..., "opinion": ..., "sentiment": ...\}]}.
\newline
\textbf{Input:} \texttt{\{sentence\}}. \\
\midrule
Few-shot
&
\textbf{Instruction:} Use the same direct extraction instruction as the
zero-shot baseline.
\newline
\textbf{Demonstrations:} \texttt{\{examples\}}, where each example consists of
an input sentence and a JSON list of gold triplets.
\newline
\textbf{Input:} \texttt{\{sentence\}}.
\newline
\textbf{Output:} return only a JSON list with the same triplet schema. \\
\midrule
CoT
&
\textbf{Instruction:} Suppose you are an expert in aspect-based sentiment
analysis. Analyze the sentence step by step: (1) identify all explicitly
mentioned aspects; (2) identify the opinion expression linked to each aspect;
(3) assign \texttt{POS}, \texttt{NEG}, or \texttt{NEU}; (4) output one triplet
for each aspect--opinion pair, including one-to-many cases; and (5) keep all
aspect and opinion spans verbatim. After brief reasoning, return the final
answer as a JSON list in the format
\texttt{[\{"aspect": ..., "opinion": ..., "sentiment": ...\}]}.
\newline
\textbf{Input:} \texttt{\{sentence\}}. \\
\midrule
CoT+few-shot
&
\textbf{Instruction:} Use the same step-by-step CoT instruction as the CoT
baseline.
\newline
\textbf{Demonstrations:} \texttt{\{cot\_examples\}}, where each example contains
an input sentence, intermediate reasoning over aspects, opinions, and sentiment
labels, and a final JSON triplet list.
\newline
\textbf{Input:} \texttt{\{sentence\}}.
\newline
\textbf{Output:} after brief reasoning, return the final JSON list with the same
triplet schema. \\
\bottomrule
\end{tabular}
\caption{Prompt templates for LLM-based baselines. The structure follows the
zero-shot, few-shot, CoT, and CoT+few-shot settings used in prior ASTE
prompting protocols, while the final output is normalized to the same JSON
triplet schema used by MASTE.}
\label{tab:baseline_prompts}
\end{table*}

\begin{table*}[!t]
\centering
\scriptsize
\setlength{\tabcolsep}{3pt}
\renewcommand{\arraystretch}{1.05}
\resizebox{\textwidth}{!}{
\begin{tabular}{lcccccccccccc}
\toprule
\multirow{2}{*}[-1.0ex]{\textbf{Ablation Setting}}
& \multicolumn{3}{c}{\textbf{14Res}}
& \multicolumn{3}{c}{\textbf{14Lap}}
& \multicolumn{3}{c}{\textbf{15Res}}
& \multicolumn{3}{c}{\textbf{16Res}} \\
\cmidrule(lr){2-4}
\cmidrule(lr){5-7}
\cmidrule(lr){8-10}
\cmidrule(lr){11-13}
& P & R & F1 & P & R & F1 & P & R & F1 & P & R & F1 \\
\midrule
MASTE (full)
& \textbf{74.64} & \textbf{70.12} & \textbf{72.31} & \textbf{64.78} & \textbf{49.24} & \textbf{55.95} & \textbf{68.95} & \textbf{66.04} & \textbf{67.46} & \textbf{74.48} & \textbf{72.35} & \textbf{73.40}\\
w/o Consistency
& 71.17 & 65.17 & 68.04 & 61.93 & 48.45 & 54.37 & 64.89 & 55.55 & 59.86 & 70.16 & 65.20 & 67.59 \\
w/o Opinion
& 65.29 & 57.06 & 60.90 & 59.04 & 45.97 & 51.69 & 65.04 & 55.20 & 59.72 & 67.58 & 61.65 & 64.48 \\
w/o Sentiment
& 66.81 & 59.32 & 62.84 & 57.86 &44.21 & 50.12 & 61.56 & 53.85 & 57.45 & 68.84 & 63.27 & 65.94 \\
w/o Opinion + Consistency
& 41.36 & 39.39 & 40.35 & 35.60 & 37.76 & 36.65 & 48.64 & 36.71 & 41.84 & 43.67 & 44.70 & 44.18 \\
\bottomrule
\end{tabular}}
\caption{Full ablation results (precision, recall, F1) with GPT-4o on four ASTE-Data-V2 benchmarks. Rows correspond to the ablation settings in Table~\ref{tab:ablation}.}
\label{tab:appendix_ablation_full}
\end{table*}

\begin{table*}[!t]
\centering
\scriptsize
\setlength{\tabcolsep}{2.8pt}
\renewcommand{\arraystretch}{1.05}
\resizebox{\textwidth}{!}{
\begin{tabular}{llcccccccccccc}
\toprule
\multirow{2}{*}{\textbf{Backbone}} & \multirow{2}{*}{\textbf{Method}}
& \multicolumn{3}{c}{\textbf{14Res}}
& \multicolumn{3}{c}{\textbf{14Lap}}
& \multicolumn{3}{c}{\textbf{15Res}}
& \multicolumn{3}{c}{\textbf{16Res}} \\
\cmidrule(lr){3-5}
\cmidrule(lr){6-8}
\cmidrule(lr){9-11}
\cmidrule(lr){12-14}
& & P & R & F1 & P & R & F1 & P & R & F1 & P & R & F1 \\
\midrule
\multirow{2}{*}{GPT-3.5-turbo}
& Zero-shot & 39.21 & 56.17 & 46.18 & 26.21 & 40.69 & 31.88 & 31.21 & 52.75 & 39.21 & 35.28 & 59.64 & 44.34 \\
& MASTE     & 62.03 & 53.42 & 57.41 & 39.52 & 33.83 & 36.45 & 54.19 & 48.04 & 50.93 & 55.04 & 57.39 & 56.19 \\
\midrule
\multirow{2}{*}{Claude Sonnet 4.6}
& Zero-shot & 57.82 & 64.69 & 61.06 & 36.87 & 40.48 & 38.59 & 49.24 & 60.00 & 54.09 & 57.79 & 69.26 & 63.01 \\
& MASTE     & 63.00 & \underline{70.22} & 66.41 & \underline{48.59} & 47.87 & 48.23 & 52.95 & 61.03 & 56.70 & \underline{61.34} & \underline{71.01} & \underline{65.83} \\
\midrule
\multirow{2}{*}{Gemini-3-flash}
& Zero-shot & 50.43 & 58.95 & 54.36 & 33.18 & 39.00 & 35.85 & 41.67 & 53.61 & 46.89 & 50.84 & 64.79 & 56.97 \\
& MASTE     & 61.23 & \textbf{75.15} & \underline{67.48} & 48.32 & \textbf{55.82} & \textbf{51.80} & \underline{54.81} & \textbf{74.02} & \textbf{62.98} & 59.97 & \textbf{77.82} & \textbf{67.74} \\
\midrule
\multirow{2}{*}{Deepseek-V3.2}
& Zero-shot & 45.85 & 49.40 & 47.55 & 29.41 & 33.27 & 31.22 & 38.14 & 46.39 & 41.86 & 46.96 & 55.64 & 50.93 \\
& MASTE     & \underline{63.26} & 69.82 & 66.38 & 48.07 & \underline{50.65} & \underline{49.32} & 53.60 & 61.44 & 57.25 & 59.73 & 69.84 & 64.39 \\
\midrule
\multirow{2}{*}{Seed 2.0 pro}
& Zero-shot & 38.90 & 39.13 & 39.02 & 26.73 & 27.91 & 27.31 & 36.43 & 29.90 & 32.84 & 41.87 & 33.07 & 36.96 \\
& MASTE & \textbf{68.20} & 67.00 & \textbf{67.60} & \textbf{49.00} & 49.33 & 49.16 & \textbf{62.50} & \underline{63.00} & \underline{62.71} & \textbf{64.80} & 66.00 & 65.32 \\

\bottomrule
\end{tabular}}
\caption{Full cross-backbone results (precision, recall, F1) across five backbone models on ASTE-Data-V2. Rows correspond to the settings in Table~\ref{tab:crossbackbone}. The best results are highlighted in bold, and the second-best results are underlined.}
\label{tab:appendix_crossbackbone_full}
\end{table*}

\FloatBarrier
\end{document}